\documentclass{article}
\usepackage{ijcai18}

\usepackage{times}
\usepackage{xcolor}
\usepackage{soul}
\usepackage[utf8]{inputenc}
\usepackage[small]{caption}
\usepackage{amsmath} 
\usepackage{amssymb} 
\DeclareMathOperator*{\argmax}{arg\,max}
\usepackage{graphicx} 
\usepackage{multirow}

\title{Spatio-Temporal Graph Convolutional Networks: A Deep Learning Framework for Traffic Forecasting}

\author{
Bing Yu \thanks{Equal contributions.}$^1$, 
Haoteng Yin$^*$$^2$$^,$$^3$, 
Zhanxing Zhu \thanks{Corresponding author.}$^3$$^,$$^4$
\\ 
$^1$ School of Mathematical Sciences, Peking University, Beijing, China\\
$^2$ Academy for Advanced Interdisciplinary Studies, Peking University, Beijing, China\\
$^3$ Center for Data Science, Peking University, Beijing, China\\
$^4$ Beijing Institute of Big Data Research (BIBDR), Beijing, China\\
\{byu, htyin, zhanxing.zhu\}@pku.edu.cn
}
\begin{document}

\maketitle

\begin{abstract}
Timely accurate traffic forecast is crucial for urban traffic control and guidance. Due to the high nonlinearity and complexity of traffic flow, traditional methods cannot satisfy the requirements of mid-and-long term prediction tasks and often neglect spatial and temporal dependencies. In this paper, we propose a novel deep learning framework, Spatio-Temporal Graph Convolutional Networks (STGCN), to tackle the time series prediction problem in traffic domain. Instead of applying regular convolutional and recurrent units, we formulate the problem on graphs and build the model with complete convolutional structures, which enable much faster training speed with fewer parameters. Experiments show that our model STGCN effectively captures comprehensive spatio-temporal correlations through modeling multi-scale traffic networks and consistently outperforms state-of-the-art baselines on various real-world traffic datasets.
\end{abstract}

\section{Introduction}
Transportation plays a vital role in everybody's daily life. According to a survey in 2015, U.S. drivers spend about 48 minutes on average behind the wheel daily.\footnote{https://aaafoundation.org/american-driving-survey-2014-2015/} Under this circumstance, accurate real-time forecast of traffic conditions is of paramount importance for road users, private sectors and governments. Widely used transportation services, such as flow control, route planning, and navigation, also rely heavily on a high-quality traffic condition evaluation. In general, multi-scale traffic forecast is the premise and foundation of urban traffic control and guidance, which is also one of main functions of the Intelligent Transportation System (ITS). 

In the traffic study, fundamental variables of traffic flow, namely speed, volume, and density are typically chosen as indicators to monitor the current status of traffic conditions and to predict the future. Based on the length of prediction, traffic forecast is generally classified into two scales: short-term (5 $\sim$ 30 min), medium and long term (over 30 min). Most prevalent statistical approaches (for example, linear regression) are able to perform well on short interval forecast. However, due to the uncertainty and complexity of traffic flow, those methods are less effective for relatively long-term predictions.

Previous studies on mid-and-long term traffic prediction can be roughly divided into two categories: dynamical modeling and data-driven methods. Dynamical modeling uses mathematical tools (e.g. differential equations) and physical knowledge to formulate traffic problems by computational simulation \cite{vlahogianni2015computational}. To achieve a steady state, the simulation process not only requires sophisticated systematic programming but also consumes massive computational power. Impractical assumptions and simplifications among the modeling also degrade the prediction accuracy. Therefore, with rapid development of traffic data collection and storage techniques, a large group of researchers are shifting their attention to data-driven approaches.

Classic statistical and machine learning models are two major representatives of data-driven methods. In time-series analysis, autoregressive integrated moving average (ARIMA) and its variants are one of the most consolidated approaches based on classical statistics \cite{ahmed1979analysis,williams2003modeling}. However, this type of model is limited by the stationary assumption of time sequences and fails to take the spatio-temporal correlation into account. Therefore, these approaches have constrained representability of highly nonlinear traffic flow. Recently, classic statistical models have been vigorously challenged by machine learning methods on traffic prediction tasks. Higher prediction accuracy and more complex data modeling can be achieved by these models, such as $k$-nearest neighbors algorithm (KNN), support vector machine (SVM), and neural networks (NN).

\textbf{Deep learning approaches} have been widely and successfully applied to various traffic tasks nowadays. Significant progress has been made in related work, for instance, deep belief network (DBN) \cite{jia2016traffic,huang2014deep}, stacked autoencoder (SAE) \cite{lv2015traffic,chen2016learning}. However, it is difficult for these dense networks to extract spatial and temporal features from the input jointly. Moreover, within narrow constraints or even complete absence of spatial attributes, the representative ability of these networks would be hindered seriously.

To take full advantage of spatial features, some researchers use convolutional neural network (CNN) to capture adjacent relations among the traffic network, along with employing recurrent neural network (RNN) on time axis. By combining long short-term memory (LSTM) network \cite{hochreiter1997long} and 1-D CNN, Wu and Tan \shortcite{wu2016short} presented a feature-level fused architecture CLTFP for short-term traffic forecast. Although it adopted a straightforward strategy, CLTFP still made the first attempt to align spatial and temporal regularities. Afterwards, Shi \emph{et al.} \shortcite{shi2015convolutional} proposed the convolutional LSTM, which is an extended fully-connected LSTM (FC-LSTM) with embedded convolutional layers. However, the normal convolutional operation applied restricts the model to only process grid structures (e.g. images, videos) rather than general domains. Meanwhile, recurrent networks for sequence learning require iterative training, which introduces error accumulation by steps. Additionally, RNN-based networks (including LSTM) are widely known to be difficult to train and computationally heavy. 

For overcoming these issues, we introduce several strategies to effectively model temporal dynamics and spatial dependencies of traffic flow. To fully utilize spatial information, we model the traffic network by a general graph instead of treating it separately (e.g. grids or segments). To handle the inherent deficiencies of recurrent networks, we employ a fully convolutional structure on time axis. Above all, we propose a novel deep learning architecture, the spatio-temporal graph convolutional networks, for traffic forecasting tasks. This architecture comprises several spatio-temporal convolutional blocks, which are a combination of graph convolutional layers \cite{defferrard2016convolutional} and convolutional sequence learning layers, to model spatial and temporal dependencies. To the best of our knowledge, it is the first time that to apply purely convolutional structures to extract spatio-temporal features simultaneously from graph-structured time series in a traffic study. We evaluate our proposed model on two real-world traffic datasets. Experiments show that our framework outperforms existing baselines in prediction tasks with multiple preset prediction lengths and network scales.

\section{Preliminary}
\subsection{Traffic Prediction on Road Graphs}
Traffic forecast is a typical time-series prediction problem, i.e. predicting the most likely traffic measurements (e.g. speed or traffic flow) in the next $H$ time steps given the previous $M$ traffic observations as, 
\begin{equation}
\begin{aligned}
\hat{v}_{t+1}&, ..., \hat{v}_{t+H}=\\
&\argmax_{v_{t+1}, ..., v_{t+H}} \text{log}~P(v_{t+1}, ..., v_{t+H}|v_{t-M+1}, ...,v_{t}),
\end{aligned}
\end{equation}
where $v_t \in \mathbb{R}^{n}$ is an observation vector of $n$ road segments at time step $t$, each element of which records historical observation for a single road segment.

\begin{figure}
	\centering
	\includegraphics[height=0.133\textwidth]{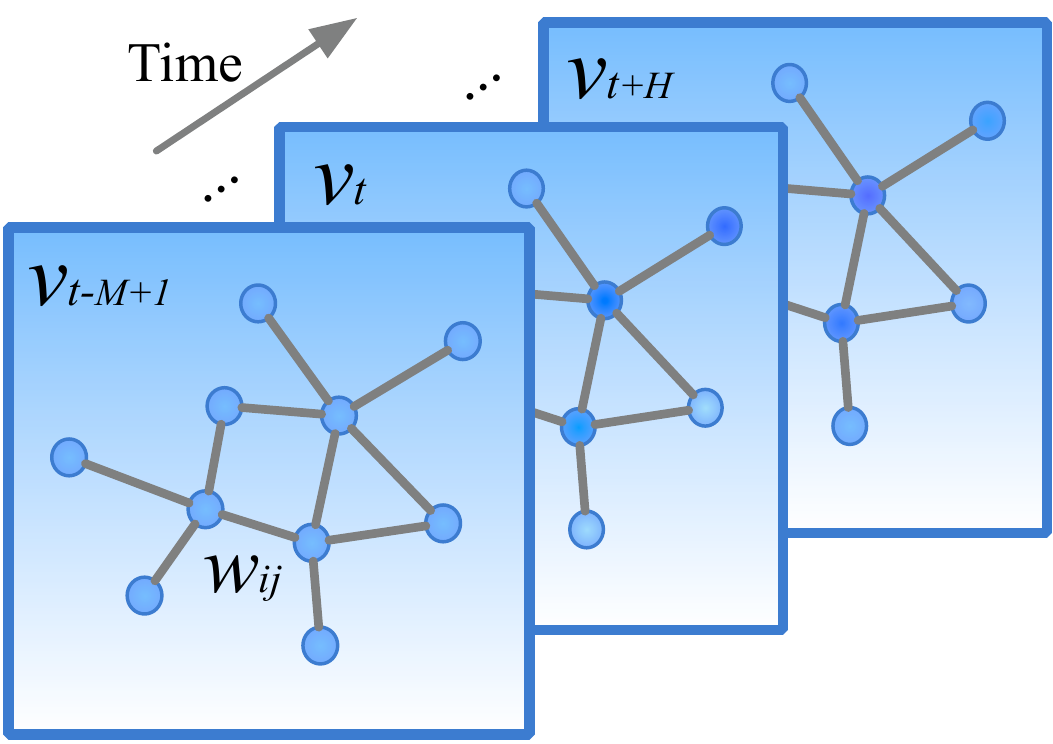}
	\caption{\label{fig:tensor}Graph-structured traffic data. Each $v_t$ indicates a frame of current traffic status at time step $t$, which is recorded in a graph-structured data matrix.}
\end{figure}

In this work, we define the traffic network on a graph and focus on structured traffic time series. The observation $v_t$ is not independent but linked by pairwise connection in graph. Therefore, the data point $v_t$ can be regarded as a graph signal that is defined on an undirected graph (or directed one) $\mathcal{G}$ with weights $w_{ij}$ as shown in Figure~\ref{fig:tensor}. At the $t$-th time step, in graph $\mathcal{G}_t=(\mathcal{V}_t,\mathcal{E}, W)$, $\mathcal{V}_t$ is a finite set of vertices, corresponding to the observations from $n$ monitor stations in a traffic network; $\mathcal{E}$ is a set of edges, indicating the connectedness between stations; while $W \in \mathbb{R}^{n \times n}$ denotes the weighted adjacency matrix of $\mathcal{G}_t$.

\subsection{Convolutions on Graphs} 
A standard convolution for regular grids is clearly not applicable to general graphs. There are two basic approaches currently exploring how to generalize CNNs to structured data forms. One is to expand the spatial definition of a convolution \cite{niepert2016learning}, and the other is to manipulate in the spectral domain with graph Fourier transforms \cite{bruna2013spectral}. The former approach rearranges the vertices into certain grid forms which can be processed by normal convolutional operations. The latter one introduces the spectral framework to apply convolutions in spectral domains, often named as the spectral graph convolution. Several following-up studies make the graph convolution more promising by reducing the computational complexity from $\mathcal{O}(n^2)$ to linear \cite{defferrard2016convolutional,kipf2016semi}.

We introduce the notion of graph convolution operator ``$*_{\mathcal{G}}$'' based on the conception of spectral graph convolution, as the multiplication of a signal $x \in \mathbb{R}^{n}$ with a kernel $\Theta$,
\begin{equation}
\Theta *_\mathcal{G} x=\Theta(L)x=\Theta(U\Lambda U^T)x=U\Theta(\Lambda)U^Tx,
\label{eq:go}
\end{equation}
where graph Fourier basis $U \in \mathbb{R}^{n \times n}$ is the matrix of eigenvectors of the normalized graph Laplacian $L=I_n - D^{-\frac{1}{2}}WD^{-\frac{1}{2}}=U \Lambda U^T \in \mathbb{R}^{n \times n}$ ($I_n$ is an identity matrix, $D \in \mathbb{R}^{n \times n}$ is the diagonal degree matrix with $D_{ii}=\Sigma_j W_{ij}$); $\Lambda \in \mathbb{R}^{n \times n}$ is the diagonal matrix of eigenvalues of $L$, and filter $\Theta(\Lambda)$ is also a diagonal matrix. By this definition, a graph signal $x$ is filtered by a kernel $\Theta$ with multiplication between $\Theta$ and graph Fourier transform $U^Tx$ \cite{shuman2013emerging}.

\section{Proposed Model}
\subsection{Network Architecture}
\begin{figure}
	\centering
	\includegraphics[width=0.48\textwidth]{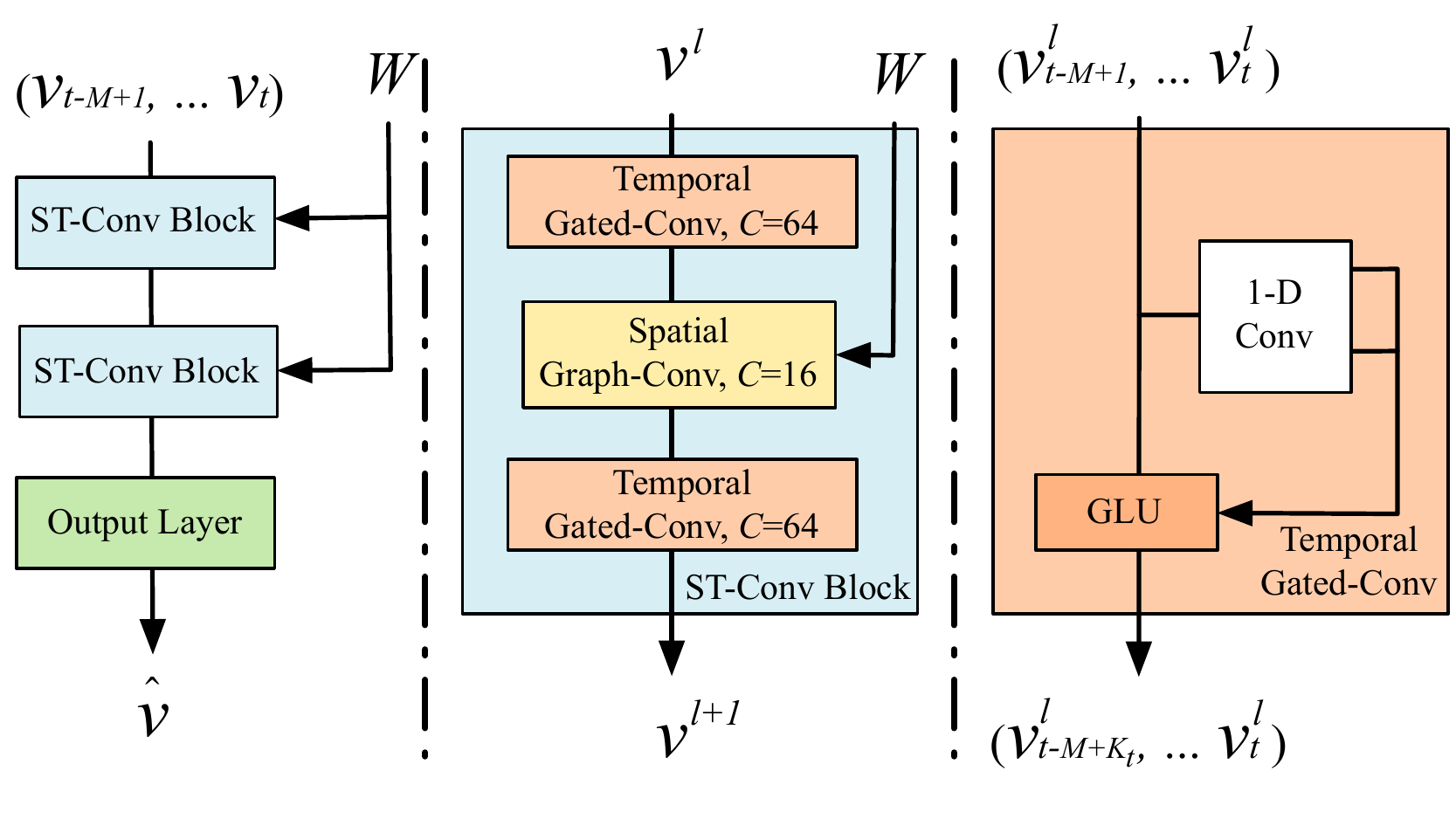}
	\caption{\label{fig:stgcn}Architecture of spatio-temporal graph convolutional networks. The framework STGCN consists of two spatio-temporal convolutional blocks (ST-Conv blocks) and a fully-connected output layer in the end. Each ST-Conv block contains two temporal gated convolution layers and one spatial graph convolution layer in the middle. The residual connection and bottleneck strategy are applied inside each block. The input $v_{t-M+1}, ..., v_{t}$ is uniformly processed by ST-Conv blocks to explore spatial and temporal dependencies coherently. Comprehensive features are integrated by an output layer to generate the final prediction $\hat{v}$.}
\end{figure}

In this section, we elaborate on the proposed architecture of spatio-temporal graph convolutional networks (STGCN). As shown in Figure~\ref{fig:stgcn}, STGCN is composed of several spatio-temporal convolutional blocks, each of which is formed as a ``sandwich'' structure with two gated sequential convolution layers and one spatial graph convolution layer in between. The details of each module are described as follows.

\subsection{Graph CNNs for Extracting Spatial Features}
The traffic network generally organizes as a graph structure. It is natural and reasonable to formulate road networks as graphs mathematically. However, previous studies neglect spatial attributes of traffic networks: the connectivity and globality of the networks are overlooked, since they are split into multiple segments or grids. Even with 2-D convolutions on grids, it can only capture the spatial locality roughly due to compromises of data modeling. Accordingly, in our model, the graph convolution is employed directly on graph-structured data to extract highly meaningful patterns and features in the space domain. Though the computation of kernel $\Theta$ in graph convolution by Eq. \eqref{eq:go} can be expensive due to $\mathcal{O}(n^2)$ multiplications with graph Fourier basis, two approximation strategies are applied to overcome this issue.

\paragraph{Chebyshev Polynomials Approximation}
To localize the filter and reduce the number of parameters, the kernel $\Theta$ can be restricted to a polynomial of $\Lambda$ as $\Theta(\Lambda) = \sum^{K-1}_{k=0} \theta_k \Lambda^k$, where $\theta \in \mathbb{R}^K$ is a vector of polynomial coefficients. $K$ is the kernel size of graph convolution, which determines the maximum radius of the convolution from central nodes. Traditionally, Chebyshev polynomial $T_k(x)$ is used to approximate kernels as a truncated expansion of order ${K\!-\!1}$ as $\Theta(\Lambda) \approx \sum^{K-1}_{k=0} \theta_k T_k(\tilde{\Lambda})$ with rescaled $\tilde{\Lambda}= 2\Lambda/\lambda_{max}-I_n$ ($\lambda_{max}$ denotes the largest eigenvalue of $L$) \cite{hammond2011wavelets}. The graph convolution can then be rewritten as,
\begin{equation}
\label{eq:goc}
\Theta *_\mathcal{G} x=\Theta(L)x \approx \sum^{K-1}_{k=0} \theta_k T_k(\tilde{L})x,
\end{equation}
where $T_k(\tilde{L}) \in \mathbb{R}^{n \times n}$ is the Chebyshev polynomial of order $k$ evaluated at the scaled Laplacian $\tilde{L}=2L/\lambda_{max}-I_n$. By recursively computing $K$-localized convolutions through the polynomial approximation, the cost of Eq. \eqref{eq:go} can be reduced to $\mathcal{O}(K|\mathcal{E}|)$ as Eq. \eqref{eq:goc} shows \cite{defferrard2016convolutional}.

\paragraph{$\text{1}^{st}$-order Approximation} 
A layer-wise linear formulation can be defined by stacking multiple localized graph convolutional layers with the first-order approximation of graph Laplacian \cite{kipf2016semi}. Consequently, a deeper architecture can be constructed to recover spatial information in depth without being limited to the explicit parameterization given by the polynomials. Due to the scaling and normalization in neural networks, we can further assume that $\lambda_{max} \approx 2$. Thus, the Eq. \eqref{eq:goc} can be simplified to, 
\begin{equation}
\begin{aligned}
\Theta *_\mathcal{G} x &\approx \theta_0 x+\theta_1 (\frac{2}{\lambda_{max}}L-I_n)x\\
&\approx \theta_0 x-\theta_1(D^{-\frac{1}{2}}WD^{-\frac{1}{2}})x,
\end{aligned}
\end{equation}
where $\theta_0$, $\theta_1$ are two shared parameters of the kernel. In order to constrain parameters and stabilize numerical performances, $\theta_0$ and $\theta_1$ are replaced by a single parameter $\theta$ by letting $\theta=\theta_0=-\theta_1$; $W$ and $D$ are renormalized by $\tilde{W}=W+I_n$ and $\tilde{D}_{ii}=\Sigma_j \tilde{W}_{ij}$ separately. Then, the graph convolution can be alternatively expressed as, 
\begin{equation}
\begin{aligned}
\Theta *_\mathcal{G} x&=\theta(I_n+D^{-\frac{1}{2}}WD^{-\frac{1}{2}})x\\
&=\theta (\tilde{D}^{-\frac{1}{2}}\tilde{W}\tilde{D}^{-\frac{1}{2}})x.
\end{aligned}
\end{equation}
Applying a stack of graph convolutions with the $\text{1}^{st}$-order approximation vertically that achieves the similar effect as $K$-localized convolutions do horizontally, all of which exploit the information from the ${(K\!-\!1)}$-order neighborhood of central nodes. In this scenario, $K$ is the number of successive filtering operations or convolutional layers in a model instead. Additionally, the layer-wise linear structure is parameter-economic and highly efficient for large-scale graphs, since the order of the approximation is limited to one.

\paragraph{Generalization of Graph Convolutions} 
The graph convolution operator ``$*_{\mathcal{G}}$'' defined on $x \in \mathbb{R}^n$ can be extended to multi-dimensional tensors. For a signal with $C_i$ channels $X \in \mathbb{R}^{n \times C_i}$, the graph convolution can be generalized by,
\begin{equation}
y_j=\sum_{i=1}^{C_i} \Theta_{i,j}(L)x_i \in \mathbb{R}^n, 1 \le j \le C_o
\end{equation}
with the $C_i \times C_o$ vectors of Chebyshev coefficients $\Theta_{i,j} \in \mathbb{R}^K$ ($C_i$, $C_o$ are the size of input and output of the feature maps, respectively). The graph convolution for 2-D variables is denoted as ``$\Theta *_\mathcal{G} X$'' with $\Theta \in \mathbb{R}^{K \times C_i \times C_o}$. Specifically, the input of traffic prediction is composed of $M$ frame of road graphs as Figure~\ref{fig:tensor} shows. Each frame $v_t$ can be regarded as a matrix whose column $i$ is the $C_i$-dimensional value of $v_t$ at the $i^{th}$ node in graph $\mathcal{G}_t$, as $X \in \mathbb{R}^{n \times C_i}$ (in this case, $C_i=1$). For each time step $t$ of $M$, the equal graph convolution operation with the same kernel $\Theta$ is imposed on $X_t \in \mathbb{R}^{n \times C_i}$ in parallel. Thus, the graph convolution can be further generalized in 3-D variables, noted as ``$\Theta *_\mathcal{G} \mathcal{X}$'' with $\mathcal{X} \in \mathbb{R}^{M \times n \times C_i}$.

\subsection{Gated CNNs for Extracting Temporal Features}
Although RNN-based models become widespread in time-series analysis, recurrent networks for traffic prediction still suffer from time-consuming iterations, complex gate mechanisms, and slow response to dynamic changes. On the contrary, CNNs have the superiority of fast training, simple structures, and no dependency constraints to previous steps. Inspired by \cite{gehring2017convolutional}, we employ entire convolutional structures on time axis to capture temporal dynamic behaviors of traffic flows. This specific design allows parallel and controllable training procedures through multi-layer convolutional structures formed as hierarchical representations.

As Figure~\ref{fig:stgcn} (right) shows, the temporal convolutional layer contains a 1-D causal convolution with a width-$K_t$ kernel followed by gated linear units (GLU) as a non-linearity. For each node in graph $\mathcal{G}$, the temporal convolution explores $K_t$ neighbors of input elements without padding which leading to shorten the length of sequences by $K_t$-1 each time. Thus, input of temporal convolution for each node can be regarded as a length-$M$ sequence with $C_i$ channels as $Y \in \mathbb{R}^{M \times C_i}$. The convolution kernel $\Gamma \in \mathbb{R}^{K_t \times C_i \times 2C_o}$ is designed to map the input $Y$ to a single output element $[P~Q] \in \mathbb{R}^{(M-K_t+1) \times (2C_o)}$ ($P$, $Q$ is split in half with the same size of channels). As a result, the temporal gated convolution can be defined as,
\begin{equation}
\Gamma *_{\mathcal{T}} Y = P \odot \sigma(Q) \in \mathbb{R}^{(M-K_t+1) \times C_o},
\end{equation} 
where $P$, $Q$ are input of gates in GLU respectively; $\odot$ denotes the element-wise Hadamard product. The sigmoid gate $\sigma(Q)$ controls which input $P$ of the current states are relevant for discovering compositional structure and dynamic variances in time series. The non-linearity gates contribute to the exploiting of the full input filed through stacked temporal layers as well. Furthermore, residual connections are implemented among stacked temporal convolutional layers. Similarly, the temporal convolution can also be generalized to 3-D variables by employing the same convolution kernel $\Gamma$ to every node $\mathcal{Y}_i \in \mathbb{R}^{M \times C_i}$ (e.g. sensor stations) in $\mathcal{G}$ equally, noted as ``$\Gamma *_{\mathcal{T}} \mathcal{Y}$'' with $\mathcal{Y} \in \mathbb{R}^{M \times n \times C_i}$.

\subsection{Spatio-temporal Convolutional Block}
In order to fuse features from both spatial and temporal domains, the spatio-temporal convolutional block (ST-Conv block) is constructed to jointly process graph-structured time series. The block itself can be stacked or extended based on the scale and complexity of particular cases. 

As illustrated in Figure~\ref{fig:stgcn} (mid), the spatial layer in the middle is to bridge two temporal layers which can achieve fast spatial-state propagation from graph convolution through temporal convolutions. The ``sandwich'' structure also helps the network sufficiently apply bottleneck strategy to achieve scale compression and feature squeezing by downscaling and upscaling of channels $C$ through the graph convolutional layer. Moreover, layer normalization is utilized within every ST-Conv block to prevent overfitting.

The input and output of ST-Conv blocks are all 3-D tensors. For the input $v^l \in \mathbb{R}^{M \times n \times C^l}$ of block $l$, the output $v^{l+1} \in \mathbb{R}^{(M-2(K_t-1)) \times n \times C^{l+1}}$ is computed by,
\begin{equation}
v^{l+1} = \Gamma^{l}_{1} *_{\mathcal{T}} \text{ReLU}(\Theta^l *_\mathcal{G} (\Gamma^{l}_{0} *_{\mathcal{T}} v^l)),
\end{equation}
where $ \Gamma_0^l$, $\Gamma_1^l$ are the upper and lower temporal kernel within block $l$, respectively; $\Theta^l$ is the spectral kernel of graph convolution; $\text{ReLU}(\cdot)$ denotes the rectified linear units function. After stacking two ST-Conv blocks, we attach an extra temporal convolution layer with a fully-connected layer as the output layer in the end (See the left of Figure~\ref{fig:stgcn}). The temporal convolution layer maps outputs of the last ST-Conv block to a single-step prediction. Then, we can obtain a final output $Z \in \mathbb{R}^{n \times c}$ from the model and calculate the speed prediction for $n$ nodes by applying a linear transformation across $c$-channels as $\hat{v}=Zw+b$, where $w \in \mathbb{R}^{c}$ is a weight vector and $b$ is a bias. We use L2 loss to measure the performance of our model. Thus, the loss function of STGCN for traffic prediction can be written as,
\begin{equation}
L(\hat{v}; W_{\theta})=\sum_t ||\hat{v}(v_{t-M+1},...,v_{t},W_{\theta})-v_{t+1}||^2,
\end{equation}
where $W_{\theta}$ are all trainable parameters in the model; $v_{t+1}$ is the ground truth and $\hat{v}(\cdot)$ denotes the model's prediction.

We now summarize the main characteristics of our model STGCN in the following, 
\begin{itemize}
	\item STGCN is a universal framework to process structured time series. It is not only able to tackle traffic network modeling and prediction issues but also to be applied to more general spatio-temporal sequence learning tasks. 
	\item The spatio-temporal block combines graph convolutions and gated temporal convolutions, which can extract the most useful spatial features and capture the most essential temporal features coherently. 
	\item The model is entirely composed of convolutional structures and therefore achieves parallelization over input with fewer parameters and faster training speed. More importantly, this economic architecture allows the model to handle large-scale networks with more efficiency.
\end{itemize}

\section{Experiments}
\subsection{Dataset Description}
We verify our model on two real-world traffic datasets, BJER4 and PeMSD7, collected by Beijing Municipal Traffic Commission and California Deportment of Transportation, respectively. Each dataset contains key attributes of traffic observations and geographic information with corresponding timestamps, as detailed below. 

\textbf{BJER4} was gathered from the major areas of east ring No.4 routes in Beijing City by double-loop detectors. There are 12 roads selected for our experiment. The traffic data are aggregated every 5 minutes. The time period used is from 1st July to 31st August, 2014 except the weekends. We select the first month of historical speed records as training set, and the rest serves as validation and test set respectively.

\begin{figure}
\begin{minipage}[t]{0.5\linewidth}
\centering
\includegraphics[height=1.35in]{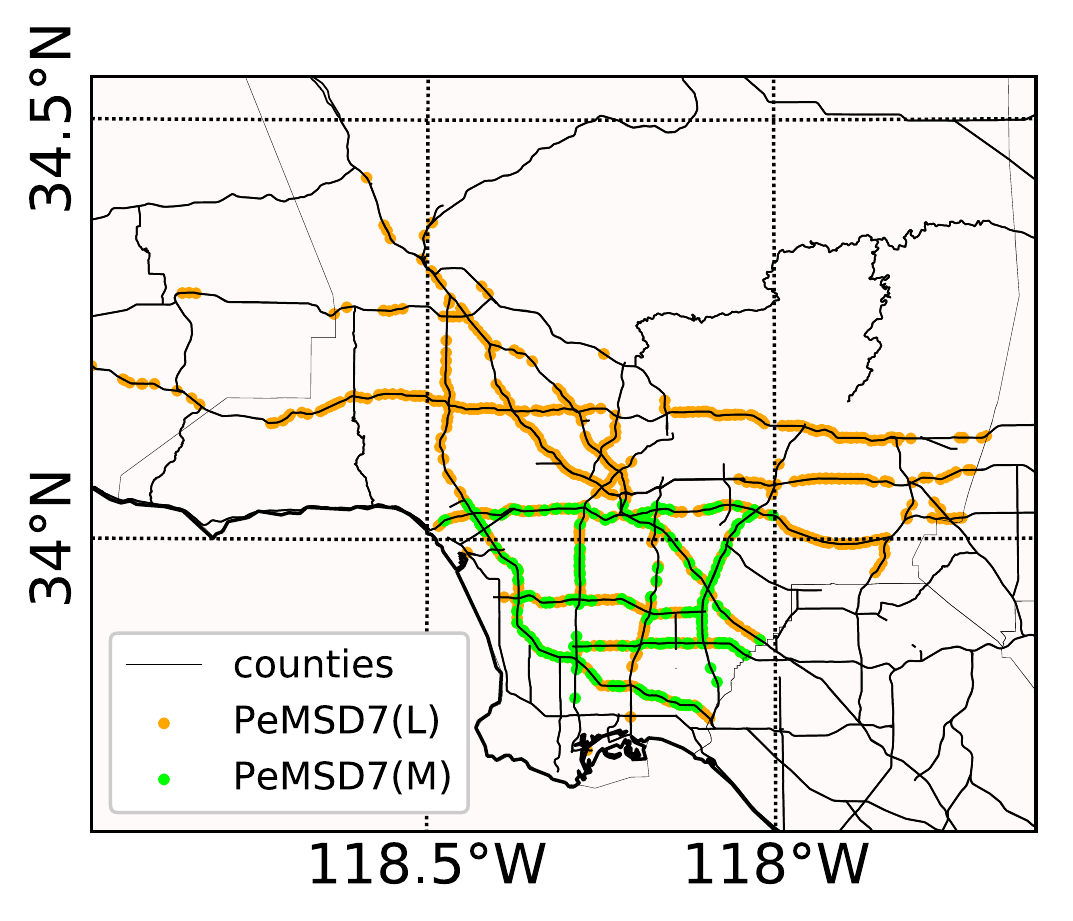}
\end{minipage}%
\begin{minipage}[t]{0.5\linewidth}
\centering
\includegraphics[height=1.35in]{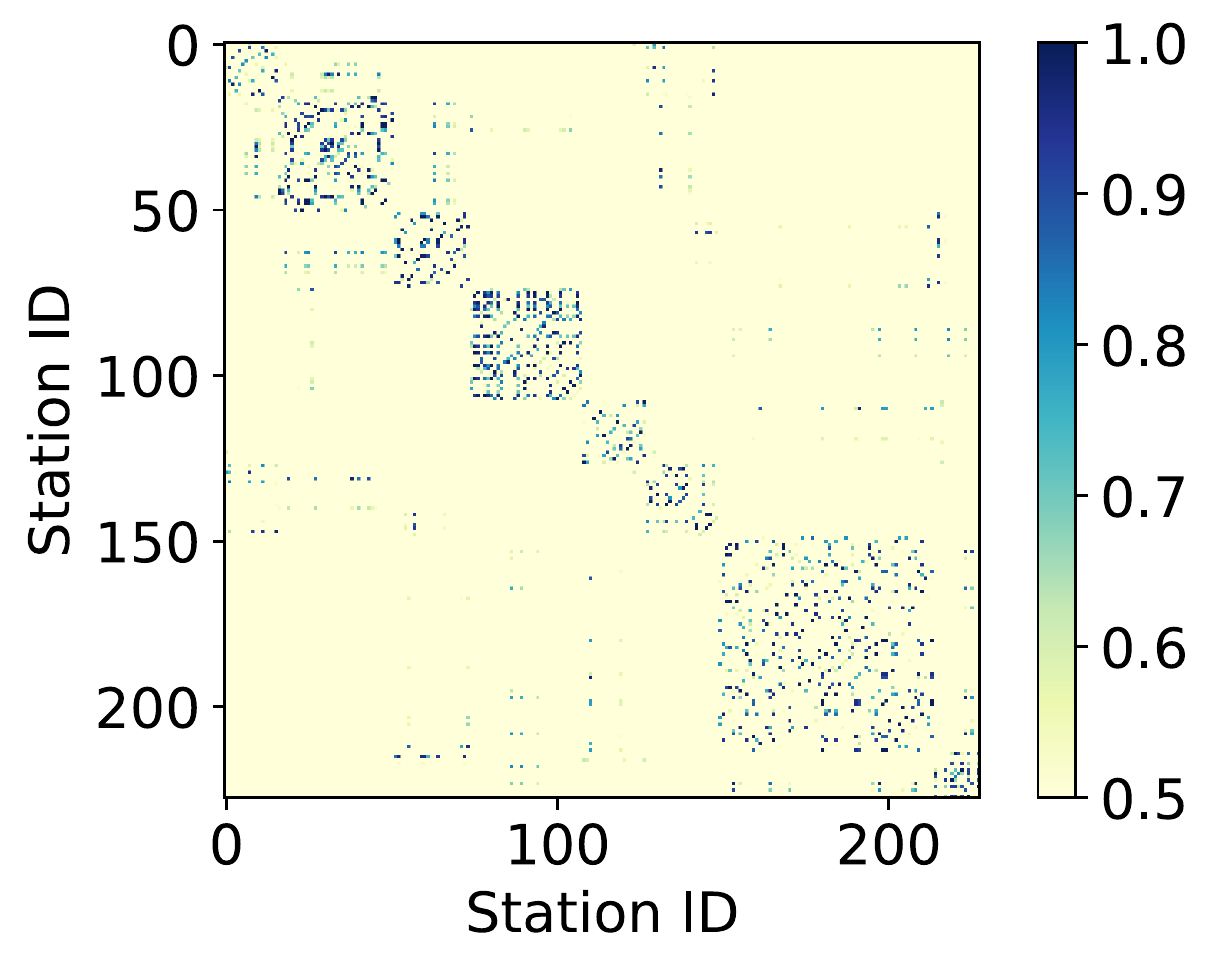}
\end{minipage}
\caption{\label{fig:pems}PeMS sensor network in District 7 of California (left), each dot denotes a sensor station; Heat map of weighted adjacency matrix in PeMSD7(M) (right).}
\end{figure}

\textbf{PeMSD7} was collected from Caltrans Performance Measurement System (PeMS) in real-time by over 39, 000 sensor stations, deployed across the major metropolitan areas of California state highway system \cite{chen2001freeway}. The dataset is also aggregated into 5-minute interval from 30-second data samples. We randomly select a medium and a large scale among the District 7 of California containing 228 and 1, 026 stations, labeled as \textbf{PeMSD7(M)} and \textbf{PeMSD7(L)}, respectively, as data sources (shown in the left of Figure~\ref{fig:pems}). The time range of PeMSD7 dataset is in the weekdays of May and June of 2012. We split the training and test sets based on the same principles as above.

\subsection{Data Preprocessing}
The standard time interval in two datasets is set to 5 minutes. Thus, every node of the road graph contains 288 data points per day. The linear interpolation method is used to fill missing values after data cleaning. In addition, data input are normalized by Z-Score method.

In BJER4, the topology of the road graph in Beijing east No.4 ring route system is constructed by the deployment diagram of sensor stations. By collating affiliation, direction and origin-destination points of each road, the ring route system can be digitized as a directed graph. 

In PeMSD7, the adjacency matrix of the road graph is computed based on the distances among stations in the traffic network. The weighted adjacency matrix $W$ can be formed as,
\begin{equation}
w_{ij}=\left\{
\begin{aligned}
	&\exp(-\frac{{d_{ij}^2}}{{\sigma^2}}),~i \neq j~\text{and}~\exp(-\frac{{d_{ij}^2}}{{\sigma^2}}) \geq \epsilon\\
	&0~~~~~~~~~~~~~~~~~,~\text{otherwise}.
\end{aligned}
\right.
\end{equation}
where $w_{ij}$ is the weight of edge which is decided by $d_{ij}$ (the distance between station $i$ and $j$). $\sigma^2$ and $\epsilon$ are thresholds to control the distribution and sparsity of matrix $W$, assigned to $10$ and $0.5$, respectively. The visualization of $W$ is presented in the right of Figure~\ref{fig:pems}.

\subsection{Experimental Settings}
All experiments are compiled and tested on a Linux cluster (CPU: Intel(R) Xeon(R) CPU E5-2620 v4 @ 2.10GHz, GPU: NVIDIA GeForce GTX 1080). In order to eliminate atypical traffic, only workday traffic data are adopted in our experiment \cite{li2015traffic}. We execute grid search strategy to locate the best parameters on validations. All the tests use 60 minutes as the historical time window, a.k.a. 12 observed data points ($M=12$) are used to forecast traffic conditions in the next 15, 30, and 45 minutes ($H=3,6,9$).

\paragraph{Evaluation Metric \& Baselines} 
To measure and evaluate the performance of different methods, Mean Absolute Errors (MAE), Mean Absolute Percentage Errors (MAPE), and Root Mean Squared Errors (RMSE) are adopted. We compare our framework STGCN with the following baselines: 
1). Historical Average (HA); 2). Linear Support Victor Regression (LSVR); 3). Auto-Regressive Integrated Moving Average (ARIMA); 4). Feed-Forward Neural Network (FNN); 5). Full-Connected LSTM (FC-LSTM) \cite{sutskever2014sequence}; 6). Graph Convolutional GRU (GCGRU) \cite{li2018dcrnn_traffic}.

\paragraph{STGCN Model} 
For BJER4 and PeMSD7(M/L), the channels of three layers in ST-Conv block are 64, 16, 64 respectively. Both the graph convolution kernel size $K$ and temporal convolution kernel size $K_t$ are set to $3$ in the model \textbf{STGCN(Cheb)} with the Chebyshev polynomials approximation, while the $K$ is set to 1 in the model \textbf{STGCN($\text{1}^{st}$)} with the $\text{1}^{st}$-order approximation. We train our models by minimizing the mean square error using RMSprop for 50 epochs with batch size as 50. The initial learning rate is $10^{-3}$ with a decay rate of 0.7 after every 5 epochs. 

\subsection{Experiment Results} 
Table \ref{tab:bjer4} and \ref{tab:pems} demonstrate the results of STGCN and baselines on the datasets BJER4 and PeMSD7(M/L). Our proposed model achieves the best performance with statistical significance (two-tailed T-test, $\alpha=0.01$, $P<0.01$) in all three evaluation metrics. We can easily observe that traditional statistical and machine learning methods may perform well for short-term forecasting, but their long-term predictions are not accurate because of error accumulation, memorization issues, and absence of spatial information. ARIMA model performs the worst due to its incapability of handling complex spatio-temporal data. Deep learning approaches generally achieved better prediction results than traditional machine learning models.

\begin{table}
\centering
\resizebox{0.48\textwidth}{!}{
\begin{tabular}{c||c|c|c}
	\hline \hline
	\multirow{2}{*}{Model} & \multicolumn{3}{|c}{BJER4 (15/ 30/ 45 min)} \\ \cline{2-4}
	& MAE & MAPE (\%) & RMSE \\ \hline \hline
	HA & 5.21 & 14.64 & 7.56 \\ \hline
	LSVR & 4.24/ 5.23/ 6.12 & 10.11/ 12.70/ 14.95 & 5.91/ 7.27/ 8.81 \\ \hline
	ARIMA & 5.99/ 6.27/ 6.70 & 15.42/ 16.36/ 17.67 & 8.19/ 8.38/ 8.72 \\ \hline
	FNN & 4.30/ 5.33/ 6.14 & 10.68/ 13.48/ 15.82 & 5.86/ 7.31/ 8.58 \\ \hline
	FC-LSTM & 4.24/ 4.74/ 5.22 & 10.78/ 12.17/ 13.60 & 5.71/ 6.62/ 7.44 \\ \hline
	GCGRU & 3.84/ 4.62/ 5.32 & 9.31/ 11.41/ 13.30 & 5.22/ 6.35/ 7.58 \\ \hline
	\textbf{STGCN(Cheb)} & \textbf{3.78}/ \textbf{4.45}/ \textbf{5.03} & \textbf{9.11}/ \textbf{10.80}/ \textbf{12.27} & \textbf{5.20}/ \textbf{6.20}/ \textbf{7.21}\\ \hline
	\textbf{STGCN($\text{1}^{st}$)} & 3.83/ 4.51/ 5.10 & 9.28/ 11.19/ 12.79 & 5.29/ 6.39/ 7.39 \\ \hline \hline
\end{tabular}}
\caption{Performance comparison of different approaches on the dataset BJER4.}
\label{tab:bjer4}
\end{table}

\begin{table*}
\centering
\resizebox{\textwidth}{!}{
\begin{tabular}{c||c|c|c||c|c|c}
	\hline \hline
	\multirow{2}{*}{Model} & \multicolumn{3}{|c||}{PeMSD7(M) (15/ 30/ 45 min)} & \multicolumn{3}{c}{PeMSD7(L) (15/ 30/ 45 min)}\\ \cline{2-7}
	& MAE & MAPE (\%) & RMSE & MAE & MAPE (\%) & RMSE \\ \hline \hline
	HA & 4.01 & 10.61 & 7.20 & 4.60 & 12.50 & 8.05 \\ \hline
	LSVR & 2.50/ 3.63/ 4.54 & 5.81/ 8.88/ 11.50 & 4.55/ 6.67/ 8.28 & 2.69/ 3.85/ 4.79 & 6.27/ 9.48/ 12.42 & 4.88/ 7.10/ 8.72 \\ \hline
	ARIMA & 5.55/ 5.86/ 6.27 & 12.92/ 13.94/ 15.20 & 9.00/ 9.13/ 9.38 & 5.50/ 5.87/ 6.30 & 12.30/ 13.54/ 14.85 & 8.63/ 8.96/ 9.39 \\ \hline
	FNN & 2.74/ 4.02/ 5.04 & 6.38/ 9.72/ 12.38 & 4.75/ 6.98/ 8.58 & 2.74/ 3.92/ 4.78 & 7.11/ 10.89/ 13.56 & 4.87/ 7.02/ 8.46 \\ \hline
	FC-LSTM & 3.57/ 3.94/ 4.16 & 8.60/ 9.55/ 10.10 & 6.20/ 7.03/ 7.51 & 4.38/ 4.51/ 4.66 & 11.10/ 11.41/ 11.69 & 7.68/ 7.94/ 8.20\\ \hline
	GCGRU & 2.37/ 3.31/ 4.01 & 5.54/ 8.06/ 9.99 & 4.21/ 5.96/ 7.13 & 2.48/ 3.43/ 4.12 ${}^{*}$ & 5.76/ 8.45/ 10.51 ${}^{*}$ & 4.40/ 6.25/ 7.49 ${}^{*}$\\ \hline
	\textbf{STGCN(Cheb)} & \textbf{2.25}/ \textbf{3.03}/ \textbf{3.57} & 5.26/ \textbf{7.33}/ \textbf{8.69} & \textbf{4.04}/ \textbf{5.70}/ \textbf{6.77} & \textbf{2.37}/ \textbf{3.27}/ \textbf{3.97} & \textbf{5.56}/ \textbf{7.98}/ \textbf{9.73} & \textbf{4.32}/ \textbf{6.21}/ \textbf{7.45}\\ \hline
	\textbf{STGCN($\text{1}^{st}$)} & 2.26/ 3.09/ 3.79 & \textbf{5.24}/ 7.39/ 9.12 & 4.07/ 5.77/ 7.03 & 2.40/ 3.31/ 4.01 & 5.63/ 8.21/ 10.12 & 4.38/ 6.43/ 7.81 \\ \hline \hline
\end{tabular}}
\caption{Performance comparison of different approaches on the dataset PeMSD7.}
\label{tab:pems}
\end{table*}

\subsubsection{Benefits of Spatial Topology} 
Previous methods did not incorporate spatial topology and modeled the time series in a coarse-grained way. Differently, through modeling spatial topology of the sensors, our model STGCN has achieved a significant improvement on short and mid-and-long term forecasting. The advantage of STGCN is more obvious on dataset PeMSD7 than BJER4, since the sensor network of PeMS is more complicated and structured (as illustrated in Figure~\ref{fig:pems}), and our model can effectively utilize spatial structure to make more accurate predictions. 

\begin{figure}
\begin{minipage}[t]{0.5\linewidth}
\centering
\includegraphics[width=1.7in]{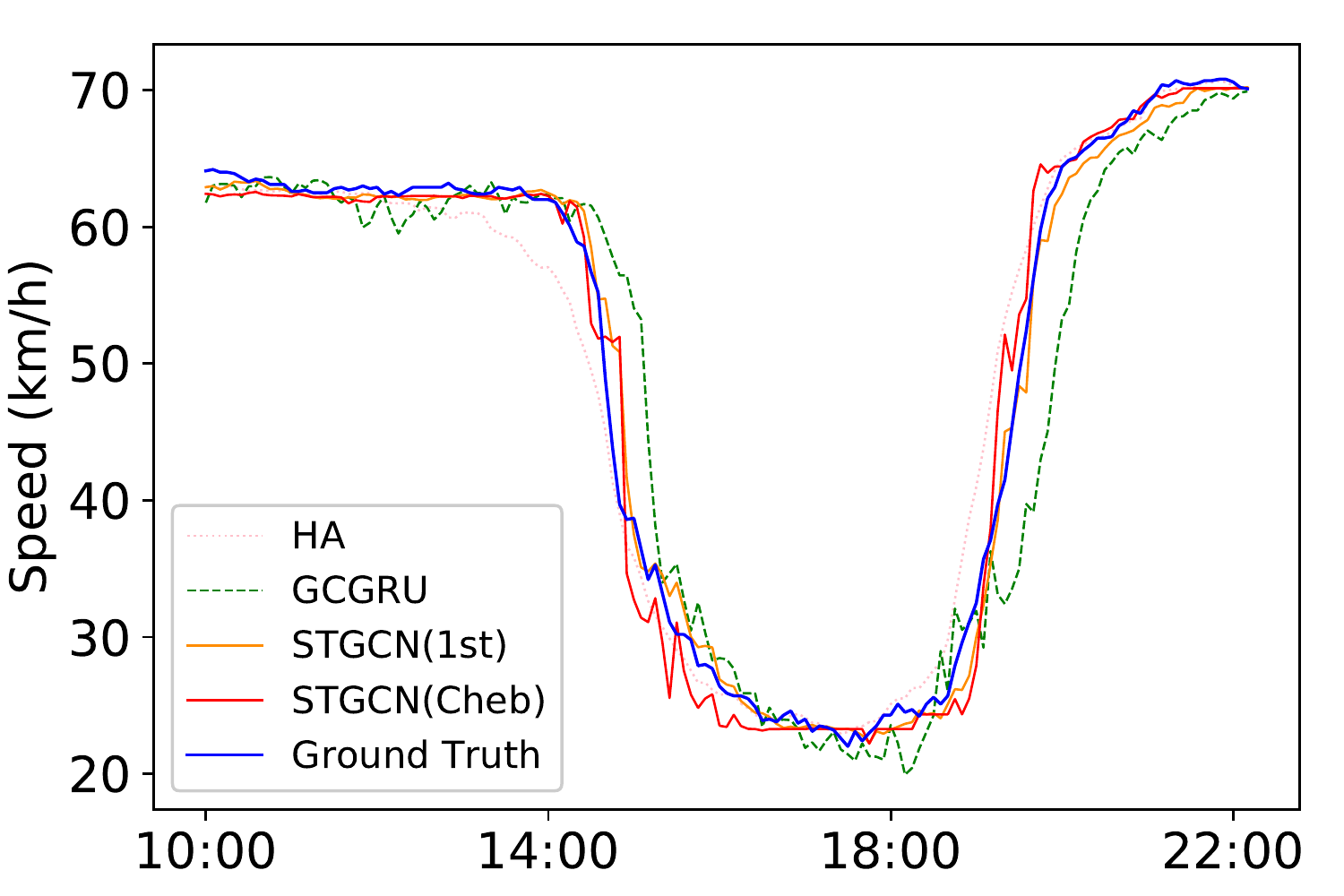}
\end{minipage}%
\begin{minipage}[t]{0.5\linewidth}
\centering
\includegraphics[width=1.7in]{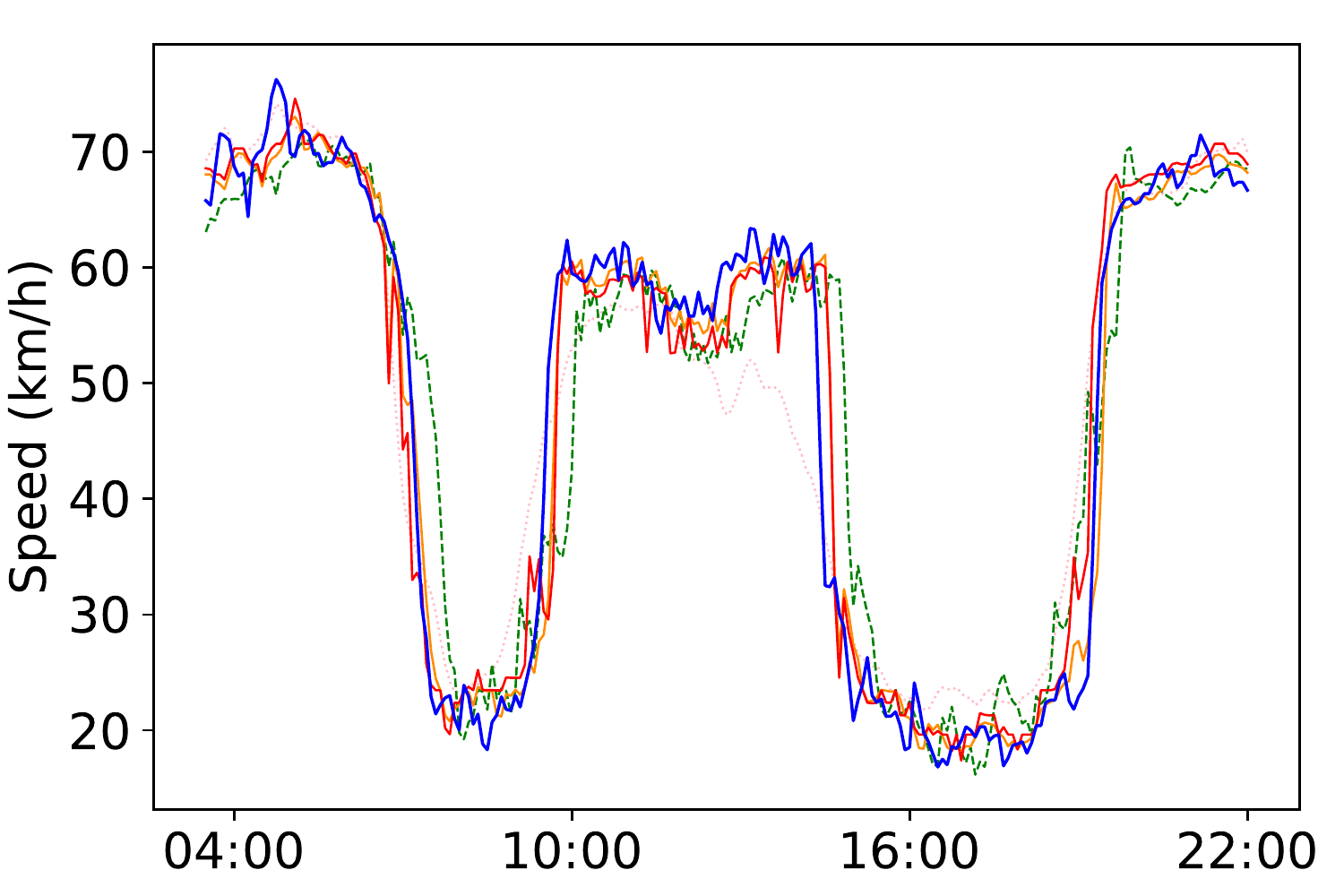}
\end{minipage}
\caption{\label{fig:sp}Speed prediction in the morning peak and evening rush hours of the dataset PeMSD7.}
\end{figure}

To compare three methods based on graph convolution: GCGRU, STGCN(Cheb) and STGCN($\text{1}^{st}$), we show their predictions during morning peak and evening rush hours, as shown in Figure~\ref{fig:sp}. It is easy to observe that our proposal STGCN captures the trend of rush hours more accurately than other methods; and it detects the ending of the rush hours earlier than others. Stemming from the efficient graph convolution and stacked temporal convolution structures, our model is capable of fast responding to the dynamic changes among the traffic network without over-reliance on historical average as most of recurrent networks do.

\begin{figure}
\begin{minipage}[t]{0.5\linewidth}
\centering
\includegraphics[width=1.7in]{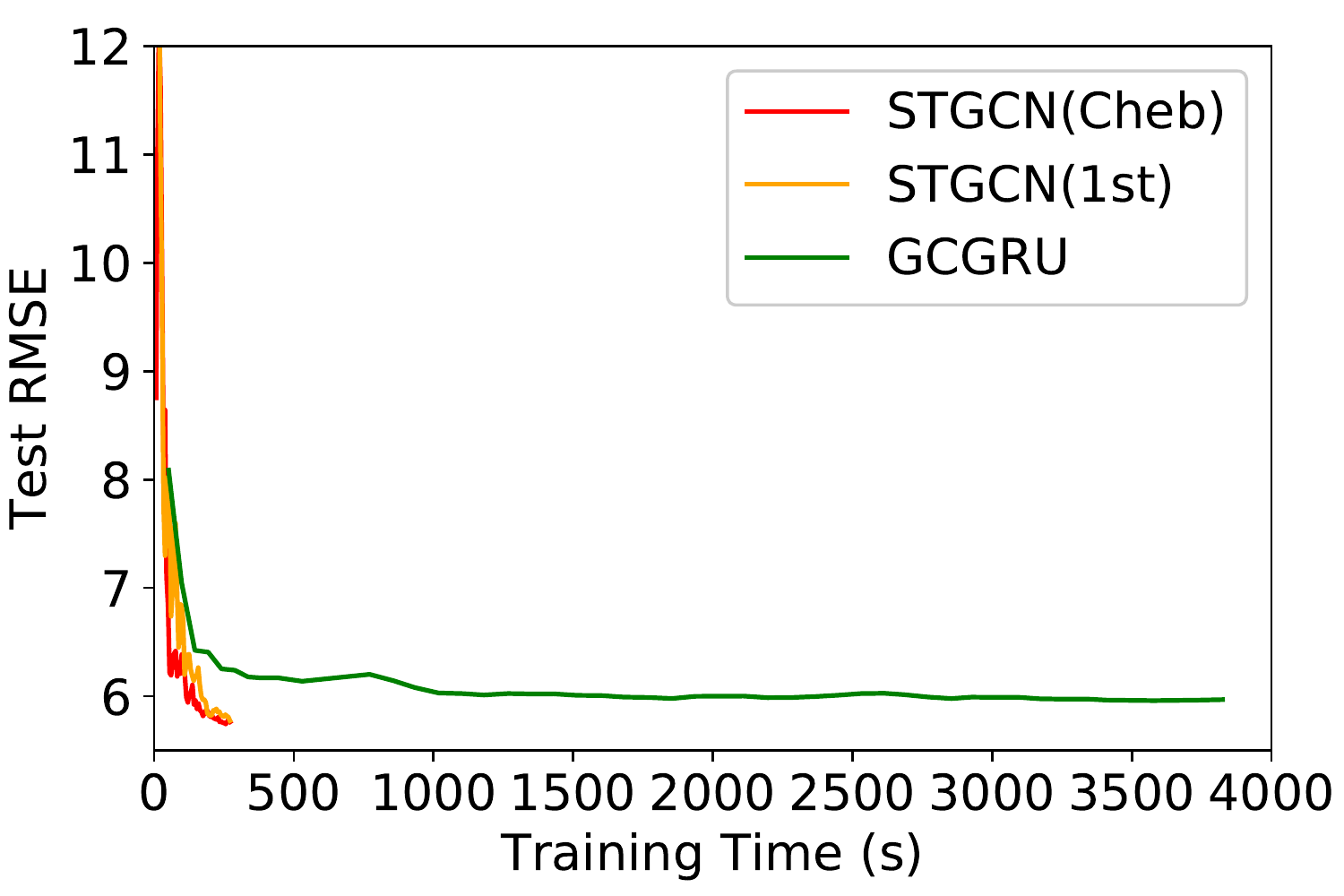}
\end{minipage}%
\begin{minipage}[t]{0.5\linewidth}
\centering
\includegraphics[width=1.7in]{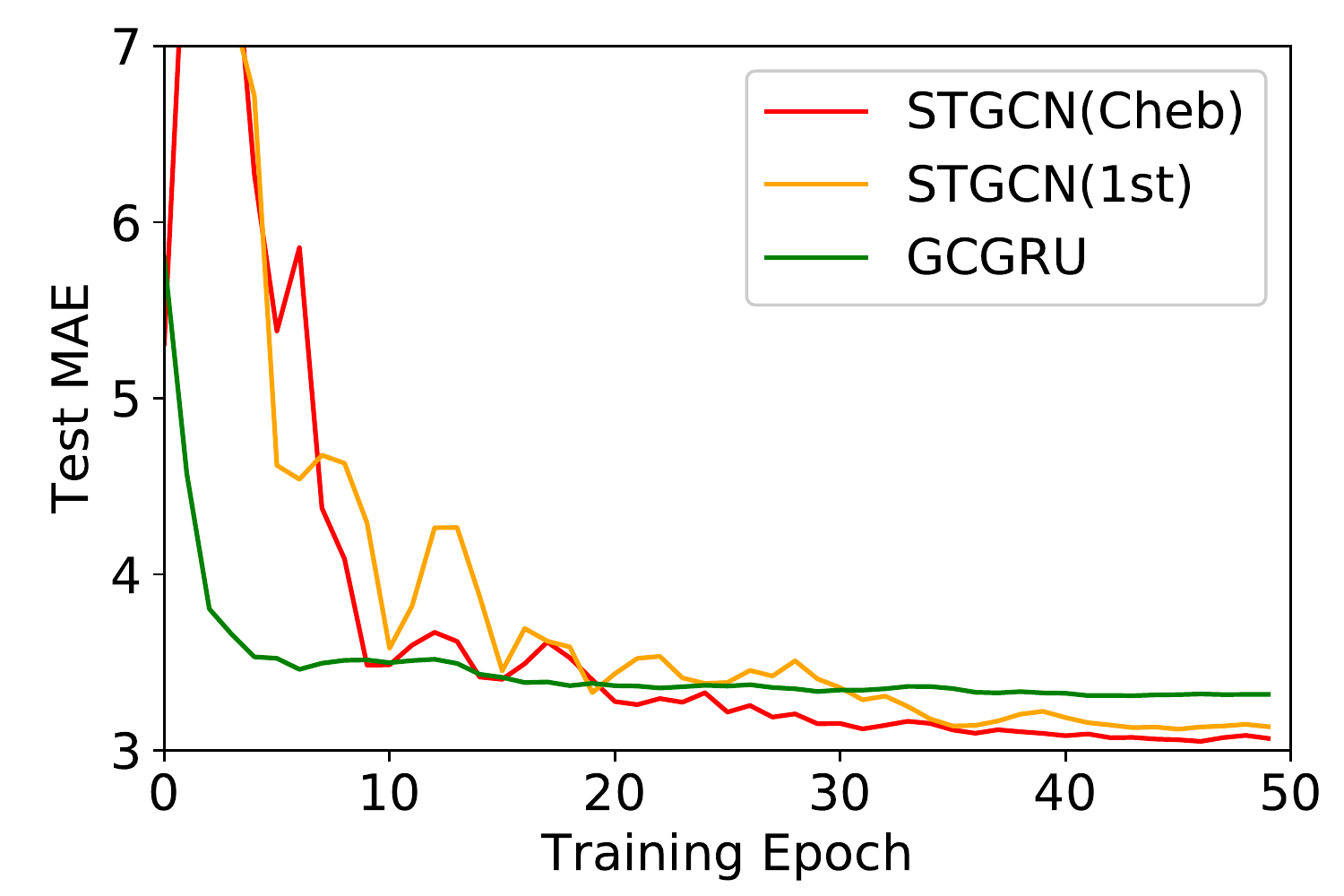}
\end{minipage}
\caption{\label{fig:rm}Test RMSE versus the training time (left); Test MAE versus the number of training epochs (right). (PeMSD7(M))}
\end{figure}

\subsubsection{Training Efficiency and Generalization} 
To see the benefits of the convolution along time axis in our proposal, we summarize the comparison of training time between STGCN and GCGRU in Table \ref{tab:tctd}. In terms of fairness, GCGRU consists of three layers with 64, 64, 128 units respectively in the experiment for PeMSD7(M), and STGCN uses the default settings as described in Section 4.3. Our model STGCN only consumes \textbf{272} seconds, while RNN-type of model GCGRU spends \textbf{3, 824} seconds on PeMSD7(M). This \textbf{14} times acceleration of training speed mainly benefits from applying the temporal convolution instead of recurrent structures, which can achieve fully parallel training rather than exclusively relying on chain structures as RNN do. For PeMSD7(L), GCGRU has to use the half of batch size since its GPU consumption exceeded the memory capacity of a single card (results marked as ``*'' in Table \ref{tab:pems}); while STGCN only need to double the channels in the middle of ST-Conv blocks. Even though our model still consumes less than a tenth of the training time of model GCGRU under this circumstance. Meanwhile, the advantages of the $\text{1}^{st}$-order approximation have appeared since it is not restricted to the parameterization of polynomials. The model STGCN($\text{1}^{st}$) speeds up around 20\% on a larger dataset with a satisfactory performance compared with STGCN(Cheb).

In order to further investigate the performance of compared deep learning models, we plot the RMSE and MAE of the test set of PeMSD7(M) during the training process, see Figure \ref{fig:rm}. Those figures also suggest that our model can achieve much faster training procedure and easier convergences. Thanks to the special designs in ST-Conv blocks, our model has superior performances in balancing time consumption and parameter settings. Specifically, the number of parameters in STGCN ($4.54 \times 10^{5}$) only accounts for around two third of GCGRU, and saving over 95\% parameters compared to FC-LSTM.

\begin{table}
\centering
\resizebox{0.48\textwidth}{!}{
\begin{tabular}{c||c|c|c}
	\hline \hline
	\multirow{2}{*}{Dataset} & \multicolumn{3}{|c}{Time Consumption (s)} \\ \cline{2-4}
	& STGCN(Cheb) & STGCN($\text{1}^{st}$) & GCGRU \\ \hline \hline
	PeMSD7(M) & \textbf{272.34} & 271.18 & 3824.54 \\ \hline
	PeMSD7(L) & 1926.81 & \textbf{1554.37} & 19511.92 \\ \hline \hline
\end{tabular}}
\caption{Time consumptions of training on the dataset PeMSD7.}
\label{tab:tctd}
\end{table}

\section{Related Works}
There are several recent deep learning studies that are also motivated by the graph convolution in spatio-temporal tasks. Seo \emph{et al.} \shortcite{seo2016structured} introduced graph convolutional recurrent network (GCRN) to identify jointly spatial structures and dynamic variation from structured sequences of data. The key challenge of this study is to determine the optimal combinations of recurrent networks and graph convolution under specific settings. Based on principles above, Li \emph{et al.} \shortcite{li2018dcrnn_traffic} successfully employed the gated recurrent units (GRU) with graph convolution for long-term traffic forecasting. In contrast to these works, we build up our model completely from convolutional structures; The ST-Conv block is specially designed to uniformly process structured data with residual connection and bottleneck strategy inside; More efficient graph convolution kernels are employed in our model as well.

\section{Conclusion and Future Work}
In this paper, we propose a novel deep learning framework STGCN for traffic prediction, integrating graph convolution and gated temporal convolution through spatio-temporal convolutional blocks. Experiments show that our model outperforms other state-of-the-art methods on two real-world datasets, indicating its great potentials on exploring spatio-temporal structures from the input. It also achieves faster training, easier convergences, and fewer parameters with flexibility and scalability. These features are quite promising and practical for scholarly development and large-scale industry deployment. In the future, we will further optimize the network structure and parameter settings. Moreover, our proposed framework can be applied into more general spatio-temporal structured sequence forecasting scenarios, such as evolving of social networks, and preference prediction in recommendation systems, etc.
\bibliographystyle{named}
\bibliography{ijcai18}

\end{document}